\documentclass[sigconf]{acmart}

\usepackage[]{algorithm2e}
\usepackage{bbm}
\usepackage{multirow}
\usepackage{subcaption}
\usepackage{float}
\usepackage{array}

\AtBeginDocument{%
  \providecommand\BibTeX{{%
    \normalfont B\kern-0.5em{\scshape i\kern-0.25em b}\kern-0.8em\TeX}}}

\copyrightyear{2019}
\acmYear{2019}
\acmDOI{}
\acmISBN{}
\acmArticle{}
\acmPrice{}

\acmConference[recsysXfashion'19]{Workshop on Recommender Systems in Fashion, 13th ACM Conference on Recommender Systems}{September 20, 2019}{Copenhagen, Denmark}

\begin{document}


\title{Automated Fashion Size Normalization}

\author{Eddie S.J. Du}
\affiliation{
  \institution{Georgian Partners}}
\email{edu@georgianpartners.com}

\author{Chang Liu}
\affiliation{
  \institution{Georgian Partners}}
\email{cliu@georgianpartners.com}

\author{David H. Wayne}
\affiliation{
  \institution{True Fit}}
\email{dwayne@truefit.com}

\begin{abstract}
The ability to accurately predict the fit of fashion items and recommend the correct size is key to reducing merchandise returns in e-commerce. A critical prerequisite of fit prediction is ``size normalization'', the mapping of product sizes across brands to a common space in which sizes can be compared. At present, size normalization is usually a time-consuming manual process. We propose a method to automate size normalization through the use of sales data. The size mappings generated from our automated approaches are comparable to human-generated mappings.

\end{abstract}



\maketitle

\section{Introduction}\label{introduction}

We are witnessing a tipping point in e-commerce as more and more people purchase goods online. Yet most clothing purchases are still made within physical stores \cite{clement_2018}. This is due to the fact that purchasing clothing and shoes online is a still a gamble for consumers. When they do shop online, many customers order multiple sizes with the purpose of returning the ones that don't fit. Not surprisingly, as online shopping for clothing and shoes grows, so have return rates. According to a recent study, 20\% of purchases made online are returned, 52\% of those indicated a problem with fit to be the reason for return \cite{orendorff}. Presenting reliable and personalized size recommendations to shoppers is a core concern for retailers. Not only will accurate recommendations reduce return rates, they will also increase engagement and boost consumer loyalty. 

True Fit is an industry leading provider of personalized size recommendations. Its fit and size recommender systems support detailed size recommendations at hundreds of different retailers with thousands of different brands. As an aggregator of retail fashion data, combining catalog and transaction data across all of its retail partners, there are unique challenges around understanding garment sizing. 

A key step to serve accurate size recommendations is understanding the wide variations in garment sizing. In the real world, the same size strings may not consistently have the same meaning. For example, the size ``small'' in a regular-size brand means a smaller fit than a ``small'' in a plus-size brand. On the other hand, sizes that look different, such as ``S'', ``SM'', ``SML'', and even ``P'' within the same brand may all mean the same fit. The relationship between different sizes, for instance ``S'' and ``6R'', is less obvious; they may or may not mean a similar fit depending on which brand each belongs to. In order to make sense of all this variation, we embed (or normalize) all the sizes across brands into a shared universal space, where sizes can be meaningfully compared with each other. We call this task ``size normalization''. In this paper, we will focus on size normalization into a 1-dimensional space.

\begin{figure}[tbhp]
  \includegraphics[width=1\linewidth]{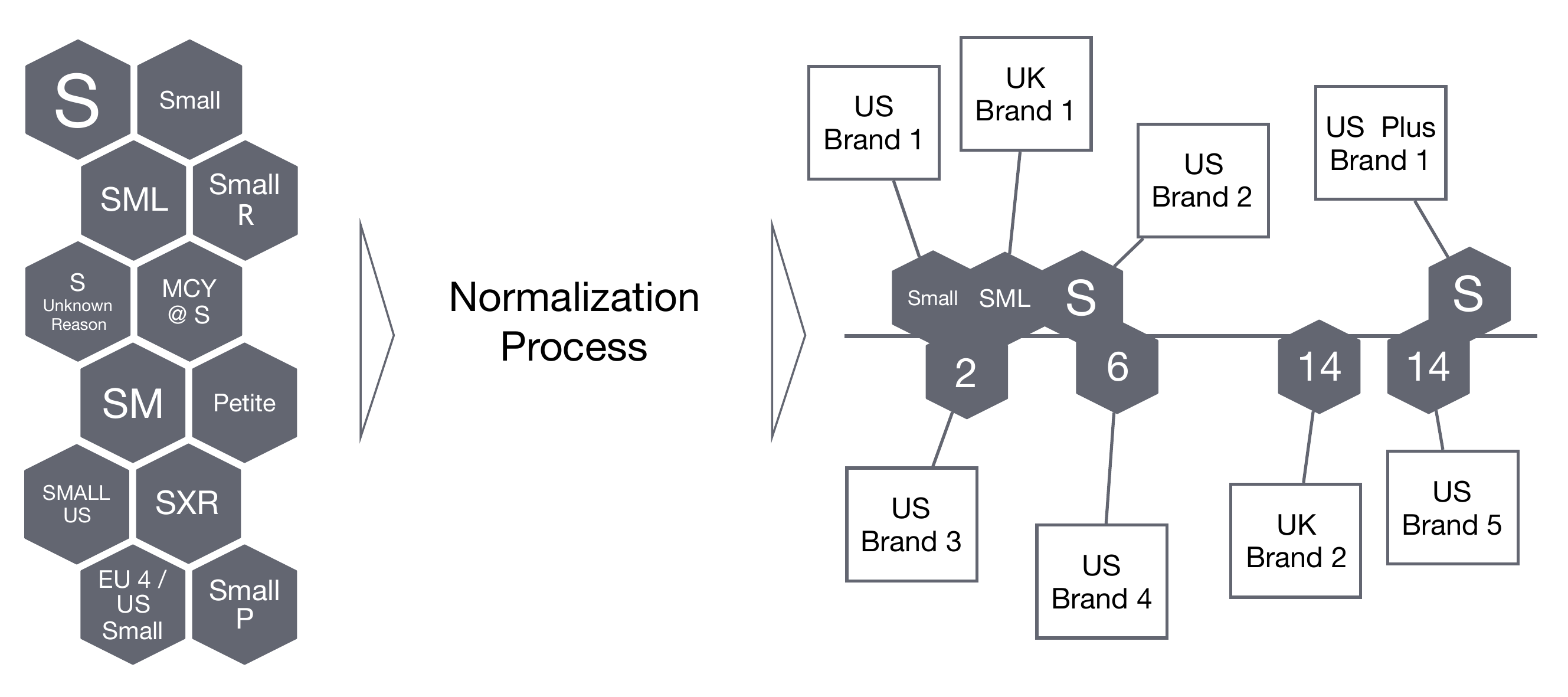}
  \caption{Sizes normalized to a universal space.}
  \label{fig:sizevariants}
\end{figure}

Traditionally, domain experts conduct size normalization by manually inspecting the sizes and the related products. This is an expensive and time-consuming process. We propose an automated size normalization framework, as shown in Figure \ref{fig:sizevariants}, using only transactions data---more specifically, data on sales where the item was not returned. We believe that size normalization systems can leverage this automated framework as part of their workflow to improve their effectiveness and efficiency.


Using sales data to normalize sizes brings two main challenges. First, connections between sizes across brands can be sparse. We rely on customers who have purchased across multiple brands to relate sizes to each other. When there is little to no customer overlap, we must rely on derived or secondary connections. Second, user buying preferences are inherently noisy due to each individual's taste. Our algorithm strives to be robust to such noise.


\smallskip
\noindent \textbf{Organization of the Paper:} We first present related research on fashion size recommendations. We then propose an automated framework to compute size normalizations strictly using sales data. Two optimization approaches are presented: a gradient descent based method and a quadratic program. Subsequently, we propose an evaluation framework for the size normalization problem and use it to compare the two optimization methods against a human-annotated size normalization approach. 

\section{Related Work} \label{lit_rev}

There is currently a variety of work that provides size recommendation. Some tools focus on electronically measuring a user's body shape from users' pictures (\cite{neophytou2013shapemate}, \cite{peng2014personalised}), while others suggest a user's body measurement by using a multiple linear regression approach and a neural network approach when given information on a user's stature, weight, span, and age \cite{lofstrom2018data}. However, in a study, authors have found that most of the users who received the correct size recommendation would not buy the size recommended due to fit preferences \cite{vecchi2015looking}.

In order to address users' individual fit preferences, many in the literature suggest leveraging users' return information and past transactions data \cite{misra2018decomposing}. One such approach uses a skip-gram based approach to size recommendation and captures the users' fit preferences by utilizing the product content data and purchase return information \cite{abdulla2017size}. The intuition is that all products purchased by a user are similar in size and fit; based on that information, the authors construct joint probability functions for products purchased by users. 
The size recommendation is then formulated as a binary classification, using the gradient boosted trees method, to predict whether a product and size will fit a specific user or not. 
In a subsequent work, the authors provide an additional graph-based approach methodology for size recommendation on shoes \cite{singh2018footwear} to combat sparsity and address the cold start problem. 
Furthermore, a group at Amazon suggests a latent factor model that predicts whether a product will fit small, right, or large to a specific customer \cite{sembium2017recommending}. The authors first present an algorithm to compute the true (latent) size of a user and a product using various loss functions. 
After computing the true sizes of users and products, a recommendation is made. This model was tested on Amazon shoe datasets. A Bayesian approach was later proposed that allowed a more robust fit probability \cite{sembium2018bayesian}. This approach was tested on the same Amazon shoe dataset and showed better results than the original non-Bayesian approach. 


In many of the proposed work described above (\cite{sembium2017recommending}, \cite{sembium2018bayesian}), the data used to compute the products' true sizes are based on clean catalog data. For example, the size ``small'' would always be spelled ``SM'', and always holds the same meaning. However, as we observe in the real world, the data comes in many different forms and often contains typos and mistakes. There seems little work in understanding and standardizing fashion products. Our work targets this specific problem of size normalization in order to provide more accurate information and inputs for size recommendation systems.




\section{Methodology} \label{methodologies}

The task of size normalization is to map each unique size in each size type to a scalar value such that sizes that offer the same fit are close together. We approach this problem in 3 steps:
\begin{enumerate}
    \item First, we group the raw size strings within each brand into brand-specific ``size types'' such as alpha sizes, numerical sizes, plus sizes, etc. by analyzing their string pattern and sorting them monotonically. 
    \item Next, we create ``frequency matrices'' that counts the co-purchases of sizes across brands and size types using the sales data.
    \item Finally, we infer a scalar value for each size in each brand-specific size types to minimize the distance between pairs of sizes that are commonly co-purchased together. 
\end{enumerate}

Note that we consider each category separately; for example, we learn a set of normalized sizes for Women's Shoes, another set for Women's Tops, another set for Men's Suits, and so on. Within each category, we consider all the brands. Size normalization is therefore useful for comparing sizes across brands within the same category.

A list of all notations used is presented in Appendix \ref{ap:notations}.

\subsection{Size Type Inference} \label{section:sizetype_inference}

A size type is unique to each brand and is defined as a set of sizes with a strict order, that is, each pair of sizes can be compared with \textit{greater than} or \textit{less than}. Specifically, sizes are compared by their semantic meanings, ie. how humans would order size strings without context. For example, sizes ``Small'' and ``Large'' from brand A can be in the same size type because ``Small'' is less than ``Large''. $[1, 2, 3, 4]$ and $[S, M, L]$ are both valid size types. $[2, 4, 6, Medium]$ is not valid, since we cannot be sure of the position of $Medium$ relative to the other sizes. Within a brand, we aim to partition all sizes into as few size types as possible. That is, while ``S,M,L'' and ``XS,XL'' are both valid size types, we prefer if they are together, ``XS,S,M,L,XL''.

Size types mainly help address data sparsity issues: typically, we only observe transactions for a few sizes in a size type; knowing the order of sizes help us infer the normalized value for the rest of the sizes. As a bonus, size types help us visualize the relationship between sizes, as seen as in Figure \ref{fig:example_freq_matrix}.

The remaining of this section describes how to partition all the sizes within a brand into size types, and how to determine the ordering of sizes within each size type.

\subsubsection{Partitioning} \label{sec:partitioning}

We propose a distance measure between size strings, then based on the distance measure, we partition all sizes available for sale within a brand into disjoint clusters. The resulting clusters are the (unordered) size types. Note we run this for each brand independently; the result is that each brand has its own set of size types.

The proposed distance measure between size strings is computed on top of string ``tokens''. The tokenization procedure works by applying regular expressions to capture substrings that are semantically meaningful (ie. sequences of numbers, sequences of characters, and punctuation) and assigning them each a token type (ie. NUMER, ALPHA, and OTHER).
For example, ``14P'' is parsed into [``14'', ``P''] with the pattern [NUMER, ALPHA]. ``12.5'' is parsed into one token, [``12.5''], with pattern [NUMER]. ``EXTRA SMALL WIDE'' is parsed into [``EXTRA SMALL'', ``WIDE''], with pattern [ALPHA, ALPHA]; as an exception, the word ``EXTRA'' followed by an alpha token is considered the same token.

Next, sizes are grouped by their token \textit{type} pattern. For example, ``14P'' with pattern [NUMER, ALPHA] is in a different group than ``SML'' with pattern [ALPHA]. A pair of sizes with different patterns have infinite distance; they definitely \textit{do not} belong to the same size type. However, sizes within the same group still may or may not belong to the same size type. For example, ``13P'' and ``13W'' both share the pattern [NUMER, ALPHA], but clearly belong to different size types. 

For each token pattern, we assume that the value at one of the positions is indicative of the size type. For example, in [``13P'', ``14P'', ``15P'', ``13W'', ``14W''], the second position has unique values of ``P'' and ``W'', which indicates two size types. Intuitively, if there less unique values at a position, it is more likely to indicate different types. Using this insight, we define $q_i$, the probability that position $i$ in a pattern of length $n$ is indicative of the size type, as follows:

\begin{equation}
\begin{split}
\hat{q}_i & := 1 - \frac{\text{total number of unique tokens in position i}}{\text{total number of unique tokens across all positions}} \\
    q_i & := \frac{e^{\beta\hat{q}_i}}{\sum_{i=1}^n e^{\beta{\hat{q}_i}}}
\end{split}
\label{eq:q_i}
\end{equation}
Here, we apply a softmax to normalize $\hat{q}_i$ into a value in a distribution. In addition, the hyperparameter $\beta$ controls how smooth the resulting distribution is. This parameter will be used later to help with the clustering step.

Let $a$ and $b$ be lists of tokens representing two size strings, both with the same token pattern of length $n$. The similarity and distance between $a$ and $b$ are defined as follow:
\begin{equation}
    sim(a, b) := \sum_{i=1}^n \mathbbm{1}[a_i = b_i] q_i 
\label{eq:sim}
\end{equation}
\begin{equation}
    dist(a, b) := 1 - sim(a, b)
\label{eq:dist}
\end{equation}

 
 



With this distance measure, any classical clustering algorithm can be employed. We use the off-the-shelf implementation of Aggolomerative Clustering with complete linkage from scikit-learn \cite{scikit-learn}. We set the number of clusters to maximize the Silhouette distance. Importantly, the Silhouette distance does not inform us when there should be only one cluster. We make this decision when the off-diagonal elements of the distance matrix has a standard deviation less than a small value $\epsilon$. In practice, we first fix $\epsilon$, then tune the value of $\beta$ to maximize the number of correct partitioning on a small hand-labeled dev set. We found $\epsilon = 0.005$ and $\beta = 15$ to work well. The resulting clusters represent different size types. 


\subsubsection{Sorting} \label{section:sorting_model}
After grouping sizes into size types, we sort the sizes using a binary classifier. With the input of two size strings, the classifier outputs 1 if the first size is semantically smaller than the second size, and 0 otherwise.

The training data for the model is taken from a limited set of size charts, with some data augmentation by randomly permuting the variations of a size string (eg. replacing ``Small'' with ``SM''). Each row of data contains a pair of sizes, $A$ and $B$, and is labelled 1 if $A$ is smaller than $B$, and 0 otherwise. After data augmentation, we had $100,000$ rows of data. We used $90,000$ for training and $10,000$ for validation.

The classification model is a 1-layer, 32-dimensional character-LSTM \cite{Hochreiter1997LongSM} followed by a fully connected layer and a sigmoid activation. In training, we concatenate the size strings (ie. into $A\_B$) then pass it to the model to predict the binary label. At inference time, we pass both $A\_B$ and $B\_A$ into the model, and whichever has a higher score determines the order. We trained with the Adam optimizer \cite{kingma2014adam} which was able to achieve 98\% validation accuracy in 30 epochs. The resulting model was reused across brands and garment types.

\subsubsection{Output} After size type inference, each brand contains its own unique set of size types. Each size in each brand is mapped to a sorted index within a size type. 

\subsection{Frequency Matrix}
We use the sales data along with the size types from the previous section to compute the frequency matrix, $F$ which counts co-purchases of sizes within each pair of size types. Let $\mathcal{B}$ be the set of unique size types, $b_i$, $b_j$ $\in \mathcal{B}$ be size type $i$ and $j$ in $\mathcal{B}$. Let $\mathcal{S}_{b_i}$ be the set of sizes in brand $b_i \in \mathcal{B}$. Then an entry in the frequency matrix $F$, $F_{(b_i, s_m), (b_j, s_n)}$ counts the number of times size $s_m$ where $m \in \mathcal{S}_{b_i}$ and size $s_n$ where $n \in \mathcal{S}_{b_j}$ are purchased together. We recognize that some users with a lot of purchases may be bulk-buyers or are buying for others, and to counter this, we dilute the count of each user by the total number of purchases that user has made. Let $\mathcal{U}$ denote the set of users, then, instead of counting 1 for each co-purchase, we count $1/P_u$ for $u \in \mathcal{U}$. The way to construct the frequency matrix is outlined in Algorithm \ref{algo:freq}.

\begin{algorithm} 
\DontPrintSemicolon
\SetAlgoLined
\KwData{Sales data and size types.}
\KwResult{$F$, a sparse frequency matrix, with default value of $0$.}
\Begin{
    $F \longleftarrow$ \text{empty sparse matrix}\;
    \For {$u \in \mathcal{U}$} {
        $P_u \longleftarrow$ all products/sizes pair user $u$ purchased and not returned \;
        \For {$(a, b) \in$ all distinct pairs in $P_u$} {
            $\text{size}_a, \text{sizetype}_a \longleftarrow$ look up size type for $a$ \;
            $\text{size}_b, \text{sizetype}_b \longleftarrow$ look up size type for $b$ \;
            $F_{(\text{sizetype}_a, \text{size}_a), (\text{sizetype}_b, \text{size}_b)} \mathrel{+}= 1/|P_u|$ \;
        }
    }
}
 \caption{Generating the Frequency Matrix. 
 }
 \label{algo:freq}
\end{algorithm}

\begin{figure}[tphb]
\includegraphics[width=\linewidth]{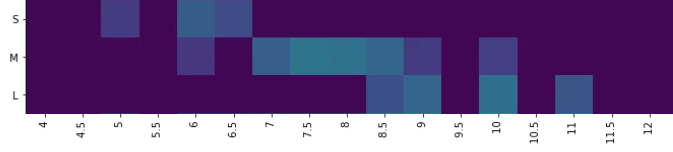}
\caption{Example of an off-diagonal block in the Frequency Matrix.}
\label{fig:example_freq_matrix}
\end{figure}

The frequency matrix is made up of block matrices, each off-diagonal block represents the relationship between a pair of size types. In Figure \ref{fig:example_freq_matrix}, we show a colour-coded example of a block matrix between two size types. The brighter the color, the higher the count. We can see that a ``S'' in one size type is around a ``5'' to ``6.5'' in the other size type, an ``M'' is around a ``6'' to ``10'', and an ``L'' is around an ``8.5'' to ``11''.
In dense blocks, we can see the relationship clearly, as shown in Figure \ref{fig:example_freq_matrix}. However, in sparser blocks, the relationship is not immediately obvious, and would need to be inferred transitively through other size types.

\subsection{Size Inference}

The frequency matrix informs us of the relationships between sizes across size types. In this step, we use those relationships to normalize sizes to a universal space. We learn a mapping of sizes that minimizes the weighted sum of squares between mapped values, where the weights are proportional to their entries in the frequency matrix. In order for the mapping to look realistic and prevent over-fitting, we also add a regularization term.
In this section, we describe the formulation in more detail, and show two implementations of the optimization procedure with quadratic programming and gradient descent.



\subsubsection{Objective Function}
The objective function that we consider here is simply the squared distance. For each pair of sizes $s_m$ and $s_n$ from size types $b_i$ and $b_j$, we compute the difference between $x_{b_i,s_m} - x_{b_j,s_n}$ and we want to minimize the total squared difference multiplied by the penalty weights from the frequency matrix as shown in Equation \ref{eq:obj}.
\begin{equation}
    \displaystyle \sum_{\substack{i=0}}^{|\mathcal{B}|} \displaystyle \sum_{\substack{j=i+1}}^{|\mathcal{B}|} \displaystyle \sum_{\substack{m=0}}^{|\mathcal{S}_{b_i}|} \displaystyle \sum_{\substack{n=0}}^{|\mathcal{S}_{b_j}|} F_{(b_i, s_m), (b_j, s_n)} * (x_{b_i,s_m} - x_{b_j,s_n})^2 \label{eq:obj} \\
\end{equation}
Furthermore, we often don't observe any transactions for sizes on the extremities, such as XXS or XXL. And so, using only the above objective function, these sizes' normalized values cannot be determined.
Therefore, we add an extra set of regularization terms to the objective functions to make sure that within each size type, the normalized sizes are placed somewhat tightly together. This allows sizes like XXS and XXL to be ``dragged along'' with the other sizes in the size type. For each size type $b_i$, we also minimize the distance between the location of the first size, $x_{b_i, s_0}$ in and the last size $x_{b_i, s_{|\mathcal{S}_{b_i}|}}$ penalized by the minimum length of the entire sizerun in size type $b_i$. The regularizer is shown in Equation (\ref{eq:reg}).
\begin{equation}
   \displaystyle \sum_{\substack{i=0}}^{|\mathcal{B}|} \frac{0.1}{|\mathcal{S}_{b_i}|} \left(x_{b_i, s_{|\mathcal{S}_{b_i}|}} - x_{b_i, s_0} \right)\label{eq:reg} \\
\end{equation}
Overall, our objective is to minimize both terms.
\begin{equation*}
    \mbox{min Equation (\ref{eq:obj})} + 
    \mbox{Equation (\ref{eq:reg})}
\end{equation*}

\subsubsection{Constraints}
We impose one set of constraints that for each size type $b_i$, the location of a larger size must be greater or equal to the closest smaller size by at least $0.1$.
    \begin{equation}
        x_{b_i,s_{m+1}} - x_{b_i,s_m} \geq 0.1 \hspace{20pt} \forall b_i \in \mathcal{B}, m \in S_{b_i} \label{cons:no}
    \end{equation}



\subsubsection{Quadratic Program (QP)}
This problem can be formulated as a quadratic program as shown in Figure \ref{QP_model}.


\begin{figure}[tbhp]
\begin{align}
    \mbox{min } & \mbox{Objective (\ref{eq:obj}) + Objective (\ref{eq:reg}}) \label{qp:obj}\\
    \mbox{s.t. } & \mbox{Constraint (\ref{cons:no}}) \notag\\
    & x_{b_i,s_m} \geq 0 & \forall b_i \in \mathcal{B}, m_i \in \mathcal{S}_{b_i} \label{c1}
\end{align}
\caption{A QP model for size normalization.}
\label{QP_model}
\end{figure}

The objective function (\ref{qp:obj}) minimizes the weighted pairwise squared difference between normalized sizes across all size types such that the location of the next size must be greater than $0.1$ than the previous size in the same size type for all the size types. Constraints (\ref{c1}) specify that all sizes must be greater than 0. Note that the $0.1$ is arbitrary and is in place to ensure separation of the different sizes.

\subsubsection{Gradient Descent (GD)}

Since we cannot enforce hard constraints with gradient descent, we need to make several adjustments. First, to satisfy the size ordering constraint (\ref{cons:no}), we introduce variables $\theta$ such that:

\begin{equation} \label{eq:gd_theta}
\begin{split}
x_{b_i,0} &= e^{\theta_{b_i,0}} \\
    x_{b_i,1} &= e^{\theta_{b_i,0}} + e^{\theta_{b_i,1}} \\
    ... \\
    x_{b_i,n} &= \sum_{k=0}^n e^{\theta_{b_i,k}}, \;\;\;\;\;\;\;\;\;\; \forall b_i \in \mathcal{B}, [0,...,n] \in \mathcal{S}_{b_i} \\
\end{split}
\end{equation}
Thereby ensuring the strictly increasing order of normalized sizes within a size type. In order to further ensure the minimum margin of $0.1$, we introduce a hinge loss:
\begin{equation} \label{eq:gd_hinge}
    \displaystyle \sum_{\substack{i=0}}^{|\mathcal{B}|} \displaystyle \sum_{\substack{m=1}}^{|\mathcal{S}_{b_i}|}
        \mbox{max} (0, x_{b_i,s_{m-1}} - x_{b_i,s_m} + 0.1) \\
\end{equation}
The complete objective we optimize is thus:
\begin{equation} \label{eq:gd_obj}
    \mbox{min Equation (\ref{eq:obj})} + 
    \alpha*\mbox{Equation (\ref{eq:reg})} + 
    \beta*\mbox{Equation (\ref{eq:gd_hinge})}
\end{equation}
In practice, we found that $\alpha=0.001$ and $\beta=100$ work well. This indicates a strong preference to ensure the minimum margin and a weak preference for sizes to stay close together. These values were tuned using another category of garments: Men's suits. Although the sizing for Men's suits is naturally different from other categories, we found that the resulting hyperparameters work well empirically.

Note that while the reparameterization to $\theta$ (Equation \ref{eq:gd_theta}) is not absolutely necessary, we found that in practice the optimization was a lot faster and more stable using it.



\section{Experiments and Results} \label{experiments}

Normalized sizes, learned with QP and GD, are compared against a set of human-annotated normalized sizes on an evaluation system described below. Human annotators were able to use any data (including size charts, product manufacturing specifications, and so on), while our method relied solely on sales data.

\subsection{Evaluation System}


With the assumption that a user's true size does not change much in a short period of time, we can expect that the sizes of that user's purchases in that period of time to be close, or ``consistent", in the normalized space. Measuring how well this holds across all users would inform to what extent we are achieving the goal of making sizes in the normalized space comparable. To do so, we propose an evaluation framework that measures the ``consistency'' of normalized sizes. The system takes as input a set of size normalization mappings and a set of test cases. Each test case is a pair of purchases, A and B, made by the same user close in time. The system looks up the normalized value of the size purchased in A, then returns the size in B with the closest normalized value. That is, the system tries to predict the size purchased in B using the size purchased in A using normalized sizes. When a size does not have a normalized value, the system abstains from making a prediction.

Two metrics are measured.
\begin{enumerate}
    \item Coverage: for how many test cases were predictions made.
    \item Accuracy: out of all the predictions made, how many of them were correct
\end{enumerate}

The definition of correctness is slightly nuanced. Variants of the same size can be normalized to exactly the same number---this happens often with human annotators. For example, let's say that ``12 Regular'' and ``12'' both map to the same normalized size, and the target answer is ``12''. In this case, either prediction should be correct, as both sizes indicate the same fit. If we assess correctness by string comparison, we would wrongly mark a correct prediction as incorrect half of the time. Instead, we defined ``correct'' to be when the \textit{human-normalized size} of the prediction and the target are equal.

\subsection{Train and Test Data}
The data we used to train and test is a two year snapshot of sales data from a subset of True Fit's cooperative of fashion retailers. Each sale contains which size was purchased, what other sizes were available at the time, and an anonymized user id. 

In total, the dataset contains 56 retailers and 5918 brands. There are approximately 60 categories ranging from Men's Tops to Unisex Kid's Shoes. The two year snapshot of sales data represents the purchases of 187 million users across 329 million orders which account for 827 million total purchased items. Across the products in this dataset, there are approximately 29 thousand distinct sizes and 150 thousand distinct product size sets. The category with the highest variation of sizes is Women's Bottoms with approximately 6,500 distinct sizes (and 16 thousand size runs). And finally the highest variation of product size sets is in the category of Women's Shoes with approximately 35 thousand distinct product size sets (comprised of groupings of approximately 5,400 women's shoe sizes).



Out of the two years of data available, we used the first year (May 2016 - Apr 2017) for training size normalization mappings. The second year (May 2017 - Apr 2018) was set aside for testing. We chose to train on a full year to reduce the effects of seasonality.

Around 400k and 300k test cases were randomly sampled for women's shoes and women's dresses respectively. Among these, 35\% and 44\% occurred in the first year (data used for training), and the rest in the second year. Each test case was generated by sampling two purchases from the same user made within the same month, and filtering out trivial scenarios (e.g. both purchases were of the same product). The same user would not be used in another test case within that month.




\subsection{Experimental Setup}

For GD, we used the Adam optimizer \cite{kingma2014adam} with learning rates of $[0.1, 0.01, 0.001]$, and trained for $40,000$ iterations with each learning rate. For QP, CPLEX 12.8 is used with default parameters and a time limit of 600 seconds. 


\subsection{Results}


First, Table \ref{table:coverage} shows the coverage in the training and test data throughout the two years. The high coverage in the first year (training set) shows that our procedure was able to assign size mappings to the vast majority of sizes used in practice. The 10\% lower coverage in the second year as compared to the first year is expected, since more brands are introduced over time. Both optimization methods, QP and GD, have the same coverage.

\begin{center}
\begin{table}[tbph]
\begin{tabular}{|l|p{3cm}|p{3cm}|}
\hline
\multicolumn{1}{|l|}{} & \multicolumn{1}{>{\centering\arraybackslash}p{3cm}|}{First Year Coverage (Training Set)}       & \multicolumn{1}{>{\centering\arraybackslash}p{3cm}|}{Second Year Coverage (Test Set)}           \\ \hline
Women's                & \multirow{2}{*}{136,081/139,164 (98\%)} & \multirow{2}{*}{225,854/254,199 (89\%)} \\ 
shoes                  &                                         &                                         \\ \hline
Women's                & \multirow{2}{*}{132,077/136,774 (97\%)} & \multirow{2}{*}{148,039/170,847 (87\%)} \\ 
dresses                &                                         &                                         \\
\hline
\end{tabular}
\caption{Coverage of automatic size normalization.}
\label{table:coverage}
\end{table}
\end{center}


Table \ref{table:accuracy} shows the accuracy of various size normalizations throughout the two years. It appears the test accuracy (accuracy in the second year) is lower than training accuracy for our automatic size normalizations.
We also include the accuracy of human-annotated size normalizations. Note that the human annotation process does not use a train-test split; sizes were normalized without transaction data. However, it does show us a benchmark of reasonable performance.
While both GD and QP are almost on par with human-annotated normalizations in the training set, the results are up to 8\% worse on the test set. This is an indication that we are perhaps over-fitting on the training data.

\begin{table}[tbph]
\begin{tabular}{|l|c|c|l|c|c|l|}
\hline
\multirow{2}{*}{} & \multicolumn{3}{>{\centering\arraybackslash}p{3cm}|}{First Year Accuracy (Training Set)} & \multicolumn{3}{>{\centering\arraybackslash}p{3cm}|}{Second Year Accuracy (Test Set)} \\ \cline{2-7} 
                  & GD         & QP          & Human       & GD        & QP        & Human      \\ \hline
Women's      & \multirow{2}{*}{62\%}       & \multirow{2}{*}{62\%}        & \multirow{2}{*}{64\%}        & \multirow{2}{*}{60\%}      & \multirow{2}{*}{60\%}      & \multirow{2}{*}{67\%}       \\ 
shoes           &                 &&&&                 &                                         \\\hline
Women's    & \multirow{2}{*}{58\%}       & \multirow{2}{*}{58\%}        & \multirow{2}{*}{59\%}        & \multirow{2}{*}{50\%}      & \multirow{2}{*}{50\%}      & \multirow{2}{*}{58\%}       \\ 
dresses                &           &&&&                  &                              \\\hline
\end{tabular}
\caption{Accuracy of automatic size normalizations compared against human-annotated mappings.}
\label{table:accuracy}
\end{table}

\begin{figure}[tphb]
    \centering
    \begin{subfigure}[b]{0.45\textwidth}
        \includegraphics[width=\textwidth]{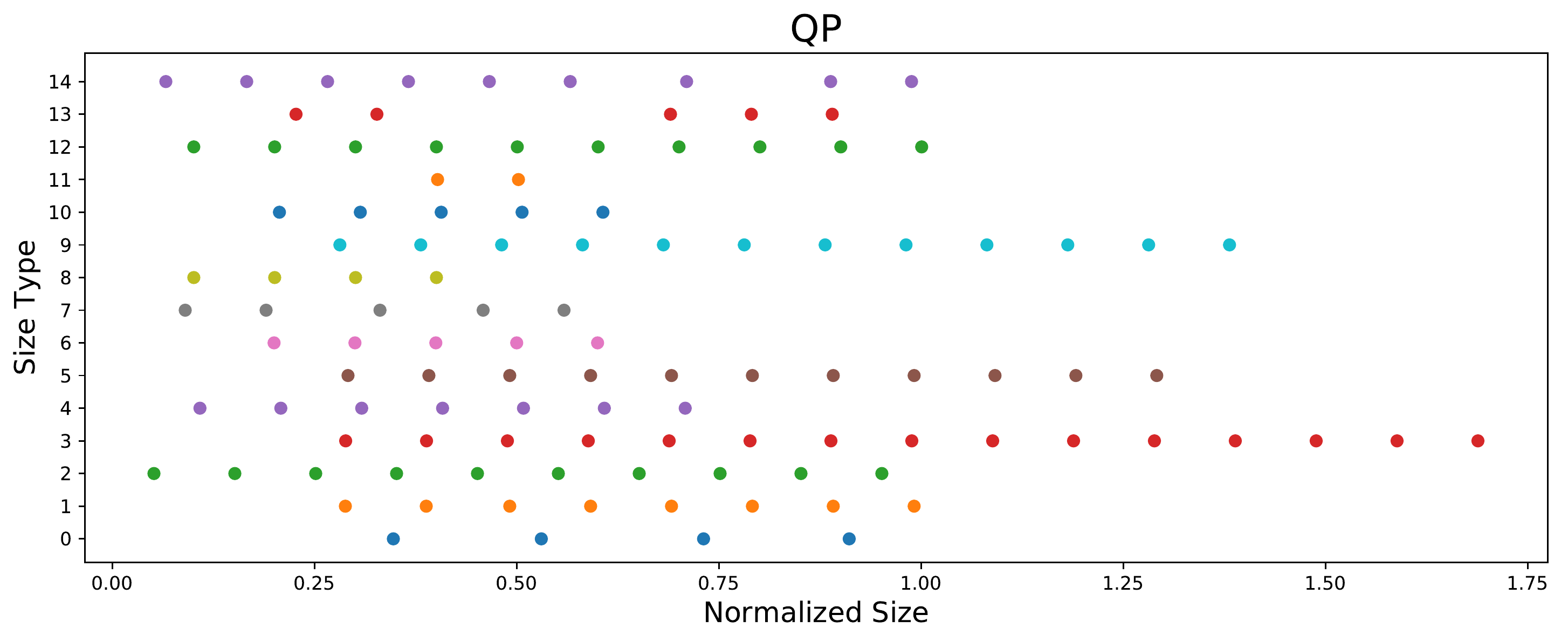}
    \end{subfigure}
    
    \begin{subfigure}[b]{0.45\textwidth}
        \includegraphics[width=\textwidth]{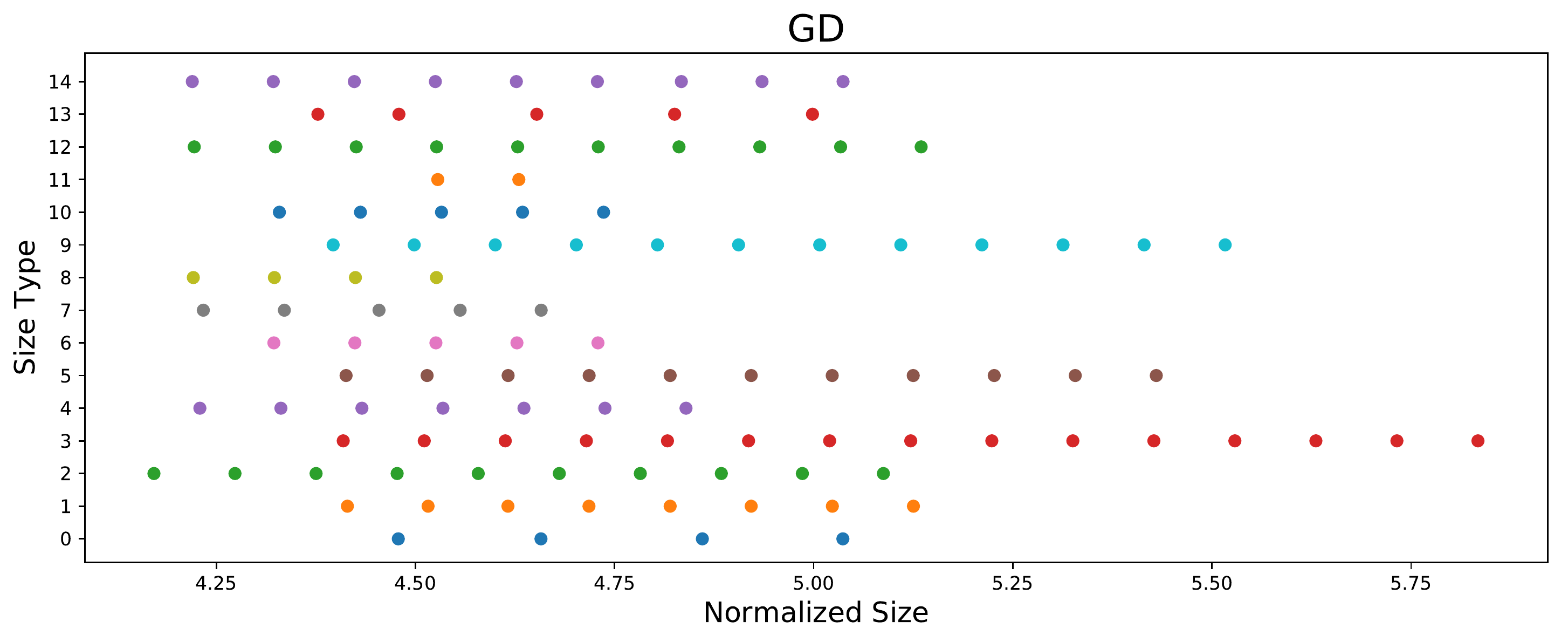}
        \caption{Women's dresses.}
    \end{subfigure}
    
    \begin{subfigure}[b]{0.45\textwidth}
        \includegraphics[width=\textwidth]{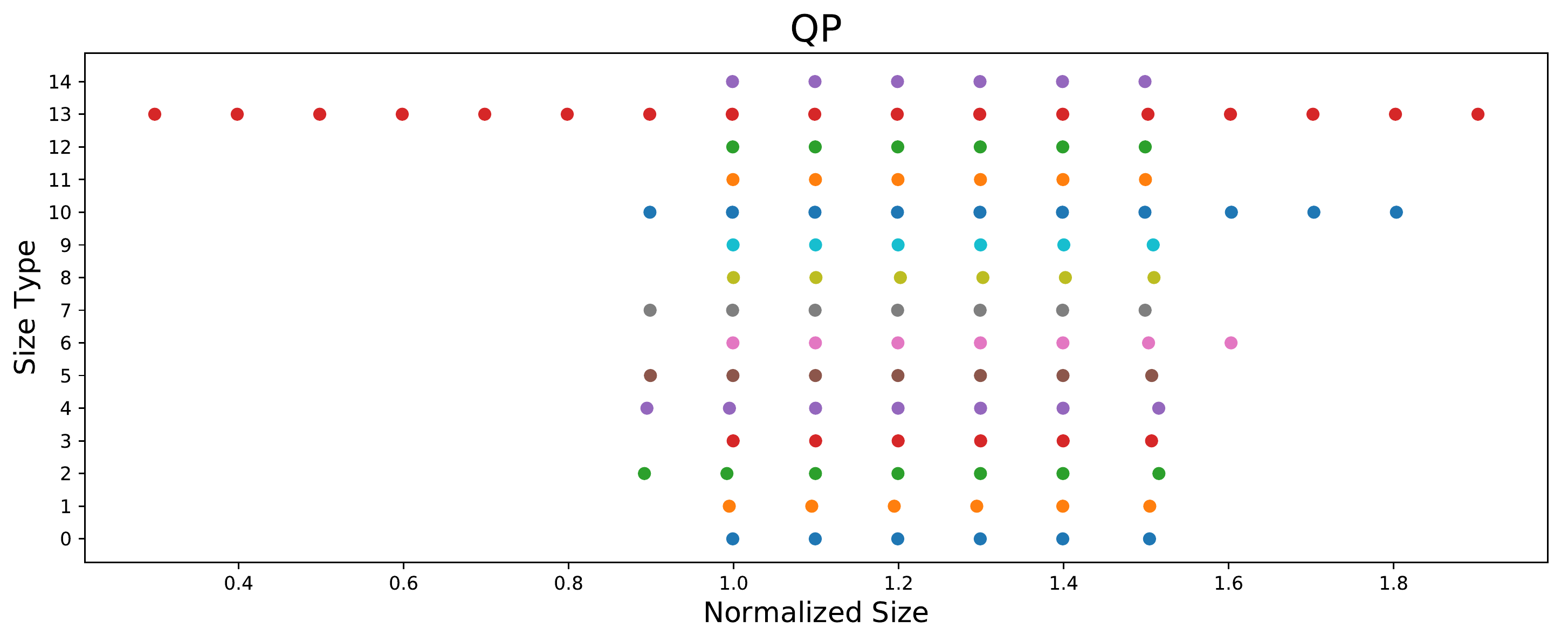}
    \end{subfigure}
    \begin{subfigure}[b]{0.45\textwidth}
        \includegraphics[width=\textwidth]{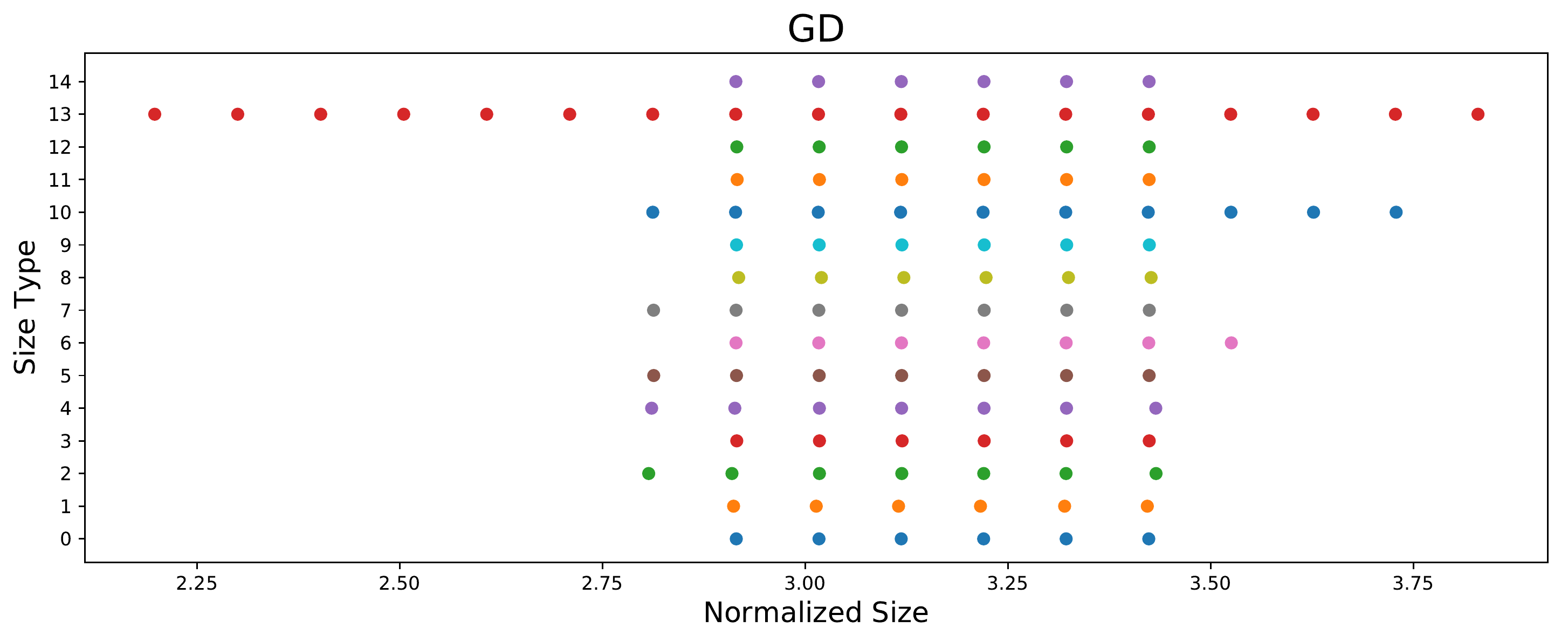}
        \caption{Women's shoes.}
    \end{subfigure}
    \caption{Normalized size mappings.}\label{fig:size_mappings}
\end{figure}

We observe that both optimization procedures, QP and GD, appear to perform equally well in terms of accuracy. Figure \ref{fig:size_mappings} shows a subsample of normalized sizes produced by GD and QP in women's dresses and women's shoes. This is expected as they are both optimizing for very similar objectives. Upon inspection, it turns out both actually produce very similar normalized sizes. However, QP has two advantages over GD. First, it is orders of magnitudes faster (Table \ref{table:runtime}). Second, it can achieve the \textit{global optimal} most of the time. We don't have the same peace of mind with GD, since we're always left wondering if the optimization could have worked better.

\begin{center}
\begin{table}[tbph]
\begin{tabular}{|l|c|c|}
\hline
                  & GD Runtime            & QP Runtime             \\ \hline
Women's shoes      & 3341s           & 33s            \\ \hline
Women's dresses    & 1206s          & 5s              \\ \hline
\end{tabular}
\caption{Approximate run-time of the two optimization methods in seconds.}
\label{table:runtime}
\end{table}
\end{center}

\section{Conclusions and Future Work}

This work explores an automated way to normalize sizes into a universal space using sales data. We introduce a fast and scalable solution and show experiments run on real-world datasets. We propose an evaluation framework for this task, and show that the automatic size normalizations perform just shy of human performance in the training set.

There are a couple of interesting opportunities for future work.
First, size type inference (Section \ref{section:sizetype_inference}) is a crucial step because any mistake there would limit the performance of everything downstream. Our proposed algorithm is static and based on heuristics. Perhaps it can be framed as a learning problem and continuously improve.
Second, since our method is completely dependant on transaction data, it is not robust when there are very few transactions. We suspect much of the drop in test accuracy may come from over-fitting on a few transactions in the training data. It would be interesting to explore how to set priors for size normalizations to account for low data scenarios. This could involve using other sources of data such as size charts, brand properties, product manufacturing specifications, and so on.
Lastly, we think it would be interesting to explore the possibility of using more than one dimension for normalized sizes. Some garments, such as dress shirts, are naturally measured by more than one dimension. Embedding all garments into a shared multi-dimensional space is very hard for humans, but should be feasible with a learned solution such as the one we propose.

\bibliographystyle{ACM-Reference-Format}
\bibliography{size_rec}


\appendix

\section{Notations} \label{ap:notations}

\begin{table}[tbhp]
\begin{tabular}{|c|p{5.5cm}|}
\hline
                    & Description                           \\ 
\hline
$\mathcal{U}$       & Set of unique users                        \\ 
$|\mathcal{U}|$     & Total number of unique users                        \\ 
$\mathcal{B}$       & Set of unique brand-sizetypes                        \\ 
$|\mathcal{B}|$     & Total number of unique brand-sizetypes                        \\
$\mathcal{S}_{b_i}$ & Set of sizes in brand $b_i \in \mathcal{B}$ \\
$|\mathcal{S}_{b_i}|$           & Total number of sizes in brand $b_i$   \\
$F_{(b_i, s_m), (b_j, s_n)}$    & Counts of how many times size $s_m$ in size type $b_i$ is purchased together with size $s_n$ in size type $b_j$\\
$x_{b_i,s_m}$       & Variable denoting the location of size $s_m$ from size type $b_i$ \\
$\theta_{b_i,s_m}$  & Auxiliary variable to compute $x_{b_i,s_m}$ in GD \\
$q_i$               & probability that position $i$ is the indicator of the size type\\
\hline
\end{tabular}
\caption{List of notations for size normalization.}
\label{table:notations}
\end{table}
\vspace{-5mm}

\section{Size Partitioning Example} \label{ap:sp_example}

A working example is shown to help provide more clarity to the algorithm described in Section \ref{sec:partitioning}. Consider we wish to partition a list of sizes into size types. Given the size strings, we first partition the sizes by regular expressions as shown in Table \ref{tab:size_partition_ex}.

\begin{table}[tbhp]
\begin{tabular}{ll}
\hline
     \textbf{Raw size strings} & \textbf{Partitioned sizes} \\
\hline
\texttt{1.5M Youth} & \texttt{['1.5', 'M', 'YOUTH']} \\
\texttt{10.5M Toddler} & \texttt{['10.5', 'M', 'TODDLER']} \\
\texttt{11.5M Toddler} & \texttt{['11.5', 'M', 'TODDLER']}\\
\texttt{11M Toddler} & \texttt{['11', 'M', 'TODDLER']}\\
\texttt{12.5M Youth} & \texttt{['12.5', 'M', 'YOUTH']}\\
\texttt{12M Toddler} & \texttt{['12', 'M', 'TODDLER']}\\
\texttt{13M Youth} & \texttt{['13', 'M', 'YOUTH']}\\
\texttt{1M Youth} & \texttt{['1', 'M', 'YOUTH']}\\
\texttt{2.5M Youth} & \texttt{['2.5', 'M', 'YOUTH']}\\
\texttt{2M Youth} & \texttt{['2', 'M', 'YOUTH']}\\
\texttt{3.5M Youth} & \texttt{['3.5', 'M', 'YOUTH']}\\
\texttt{3.5W Youth} & \texttt{['3.5', 'W', 'YOUTH']}\\
\texttt{3M Youth} & \texttt{['3', 'M', 'YOUTH']}\\
\texttt{4.5W Youth} & \texttt{['4.5', 'W', 'YOUTH']}\\
\texttt{4M Youth} & \texttt{['4', 'M', 'YOUTH']}\\
\texttt{4W Youth} & \texttt{['4', 'W', 'YOUTH']}\\
\texttt{5.5W Youth} & \texttt{['5.5', 'W', 'YOUTH']}\\
\texttt{5M Youth} & \texttt{['5', 'M', 'YOUTH']}\\
\texttt{5W Youth} & \texttt{['5', 'W', 'YOUTH']}\\
\texttt{6.5W Youth} & \texttt{['6.5', 'W', 'YOUTH']}\\
\texttt{6M Youth} & \texttt{['6', 'M', 'YOUTH']}\\
\texttt{6W Youth} & \texttt{['6', 'W', 'YOUTH']}\\
\texttt{7W Youth} & \texttt{['7', 'W', 'YOUTH']}\\
\hline
\end{tabular}
\caption{A size partition example.}
\label{tab:size_partition_ex}
\end{table}
\vspace{-5mm}

In this example, all sizes have the same pattern, [NUMER, ALPHA, ALPHA]. There are 19, 2, and 2 unique tokens in each position respectively, for a total of 23 unique tokens in total. We use this information to compute $\hat{q}$:
$$
\hat{q} = [0.17, 0.91, 0.91]
$$

We then pass $\hat{q}$ through a softmax with $\beta=15$. The softmax function normalizes $\hat{q}$ into a distribution, and the parameter $\beta$ makes the values more polarized. Note that more polarity effectively makes points that are closer to be even closer, and points further apart to be even more further apart. Therefore, finding the right amount of polarity helps to determine the right number of clusters. This is why we opt to fix the method to find number of clusters, then tune the $\beta$ parameter until we reach a value that can accurately determine the number of clusters on a development set. The result of softmax is:
$$
q = [0, 0.5, 0.5]
$$

\begin{figure}[tbhp]
    \begin{subfigure}[b]{0.47\columnwidth}
         \includegraphics[width=\linewidth]{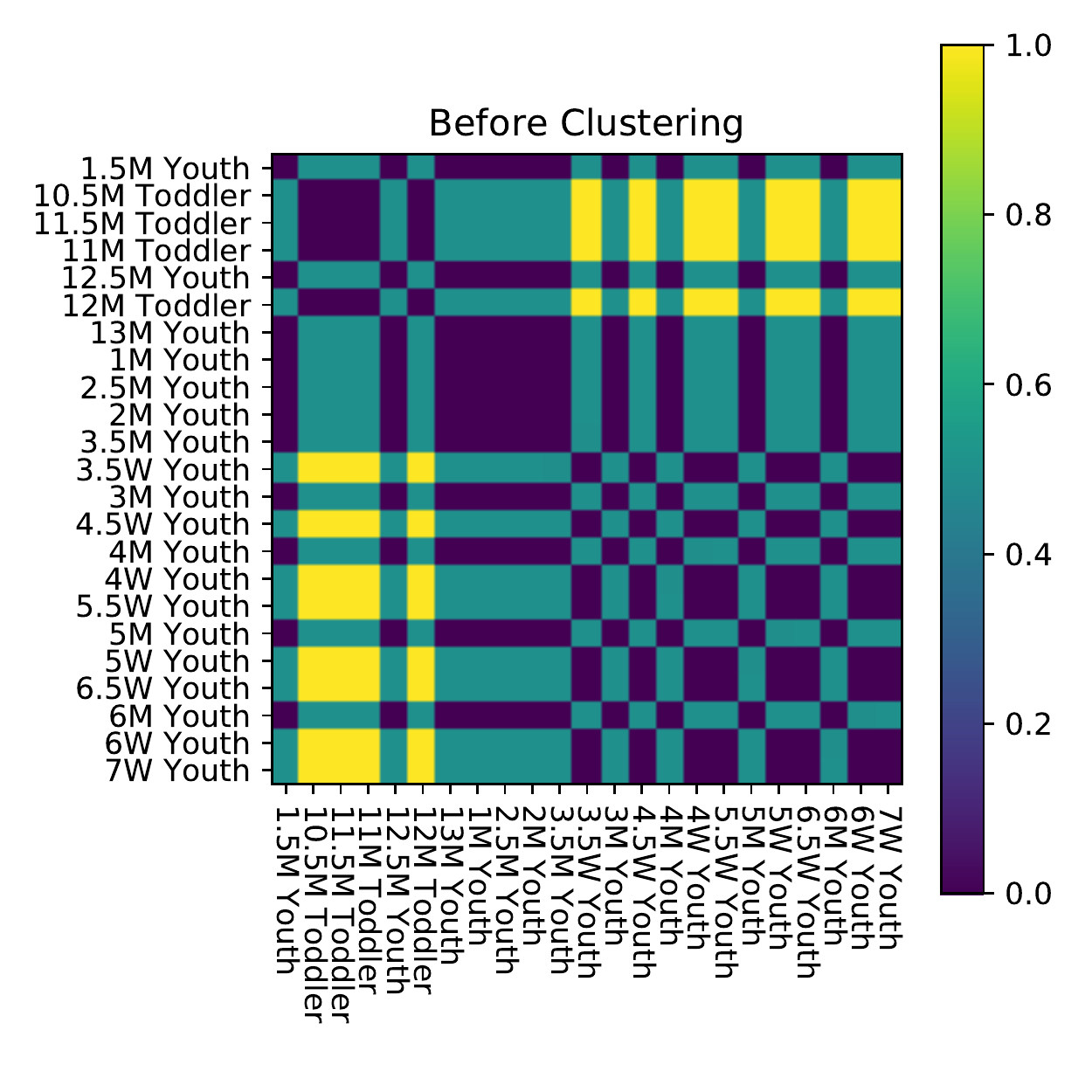}
        \caption{Distance matrix before clustering.}
        \label{fig:before_clustering}
    \end{subfigure}
     \hfill 
    \begin{subfigure}[b]{0.47\columnwidth}
        \includegraphics[width=\linewidth]{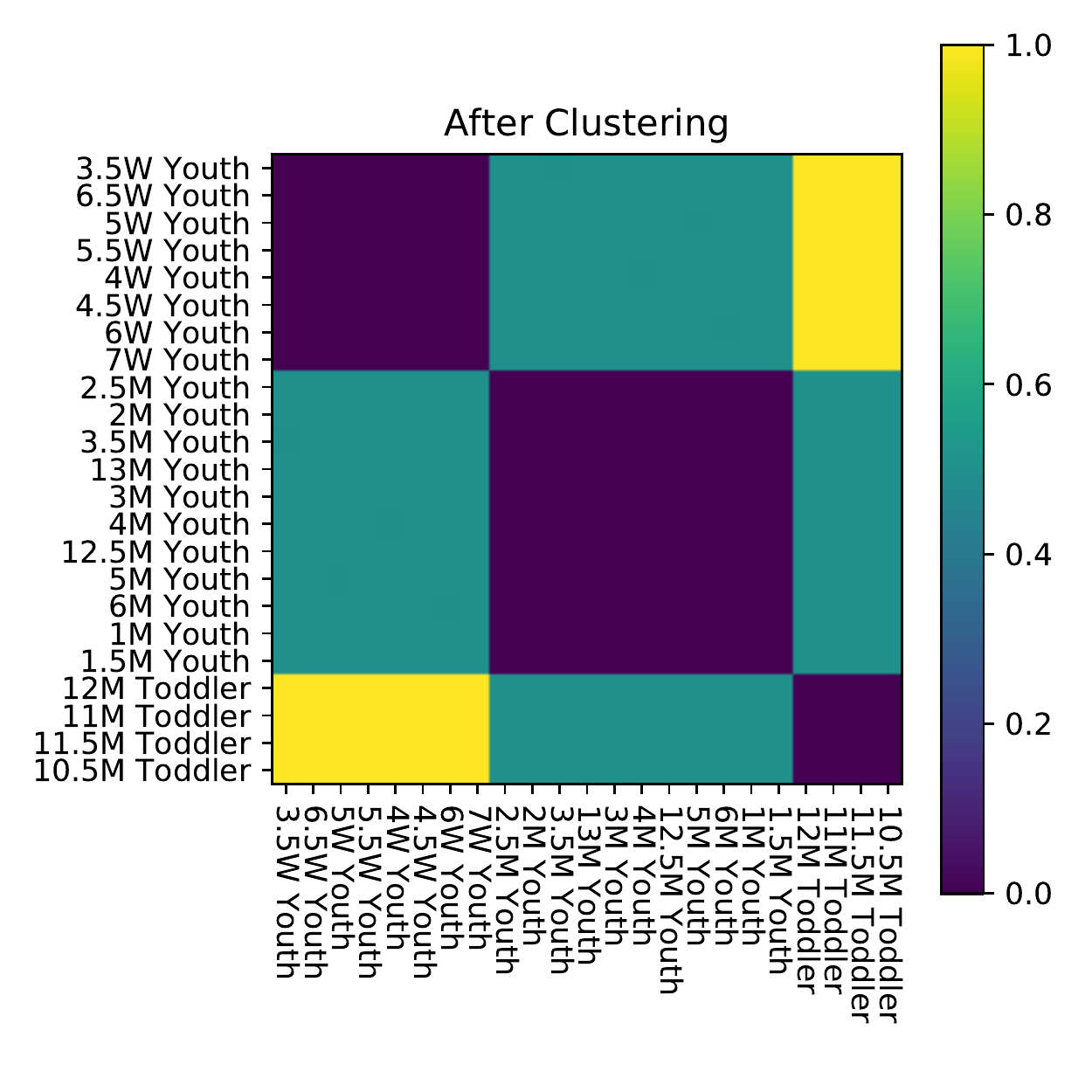}
        \caption{Distance matrix after clustering.}
        \label{fig:after_clustering}
    \end{subfigure}
    \label{fig:clusters}
    \caption{Example distance matrices.}
\end{figure}

\begin{figure}[tbhp]
    \includegraphics[width=0.45\textwidth, height = 0.88\textheight, keepaspectratio]{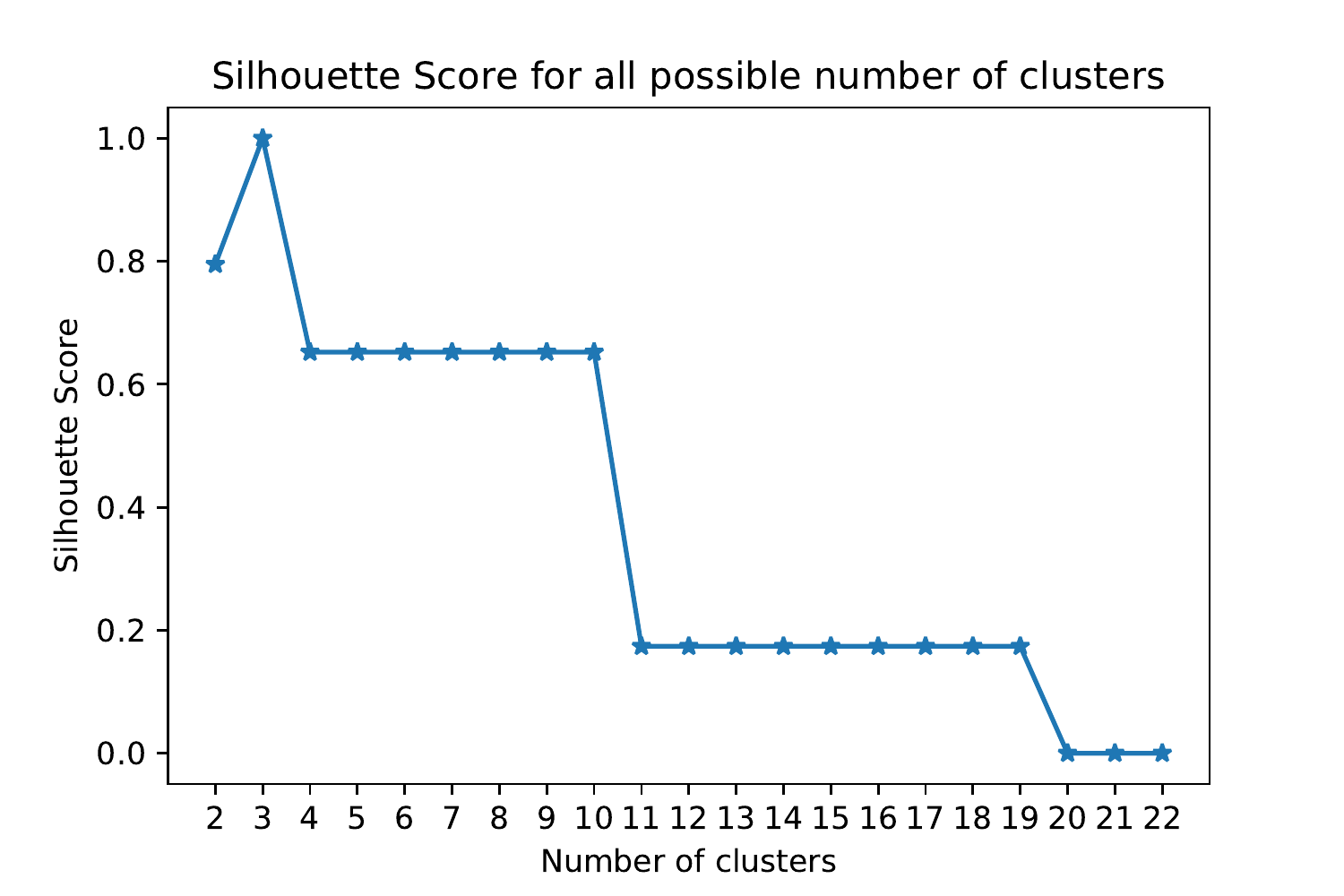}
    \caption{Silhouette Score of different number of clusters.}
    \label{fig:sil}
\end{figure}

Next, Equation \ref{eq:dist} is used to compute the distance between all pairs of sizes. This resulting distance matrix is shown in Figure \ref{fig:before_clustering}. The Silhouette Score is computed on all possible number of clusters, see Figure \ref{fig:sil}. In this case, it appears that 3 clusters is optimal. Finally, we run Hierarchical Clustering with the aim to find 3 clusters. This results in 3 size types, as one can see in Figure \ref{fig:after_clustering}.

\vspace{-3mm}

\vspace{-3mm}

A reader who understands US kids shoe sizing might notice that the ``M'' in the toddler size represents ``months'', while the ``M'' in the youth size represents ``medium''. Our proposed method gets around the need to assign such meaning to sizes while still achieving semantically meaningful partitions most of the time.



\end{document}